\documentclass[10pt,twocolumn,letterpaper]{article}

\usepackage{cvpr}      

\definecolor{cvprblue}{rgb}{0.21,0.49,0.74}
\usepackage[pagebackref,breaklinks,colorlinks,allcolors=cvprblue]{hyperref}

\def\paperID{*} 
\def\confName{CVPR}
\def\confYear{2025}

\title{What’s in a Video: Factorized Autoregressive Decoding for \\Online Dense Video Captioning}

\author{AJ Piergiovanni \\
\and
Dahun Kim
\and
Michael S. Ryoo \\
 \\
Google Deepmind
\and
Isaac Noble
\and
Anelia Angelova
}

\begin{document}
\maketitle

\begin{abstract}


Generating automatic dense captions for videos that accurately describe their contents remains a challenging area of research. Most current models require processing the entire video at once. Instead, we propose an efficient, online approach which outputs frequent, detailed and temporally aligned captions, without access to future frames. Our model uses a novel  
autoregressive factorized decoding architecture, which models the sequence of visual features for each time segment, outputting localized descriptions and efficiently leverages the context from the previous video segments. This allows the model to output frequent, detailed captions to more comprehensively describe the video, according to its actual local content, rather than mimic the training data. Second, we propose an optimization for efficient training and inference, which enables scaling to longer videos. Our approach shows excellent performance compared to both offline and online methods, and uses 20\% less compute. The annotations produced are much more comprehensive and frequent, and can further be utilized in automatic video tagging and in large-scale video data harvesting.

\end{abstract}







\section{Introduction}

With the rapid growth of video content, the need to automatically understand and annotate videos is becoming even more important~\cite{ucf101,Carreira_2017_CVPR,kay_arxiv_2017,somethingsomething,monfort_pami_2019}. Despite progress in action recognition, video retrieval, video question answering, video captioning~\cite{Carreira_2017_CVPR,mPLUG-2,actionformer,merlot,videoOFA,InternVideo,UniVL,li2023umt,wang2022git,fu2021violet}, 
approaches addressing these major video tasks are mostly focused on a single important event or activity per video. 
However, videos contain rich information with a variety of events or actions in a sequence. Thus, in order to fully capture the contents of a video, one needs many dense captions per video, rather than a single, global caption.

\begin{figure}
    \centering
   \includegraphics[width=1.0\linewidth]{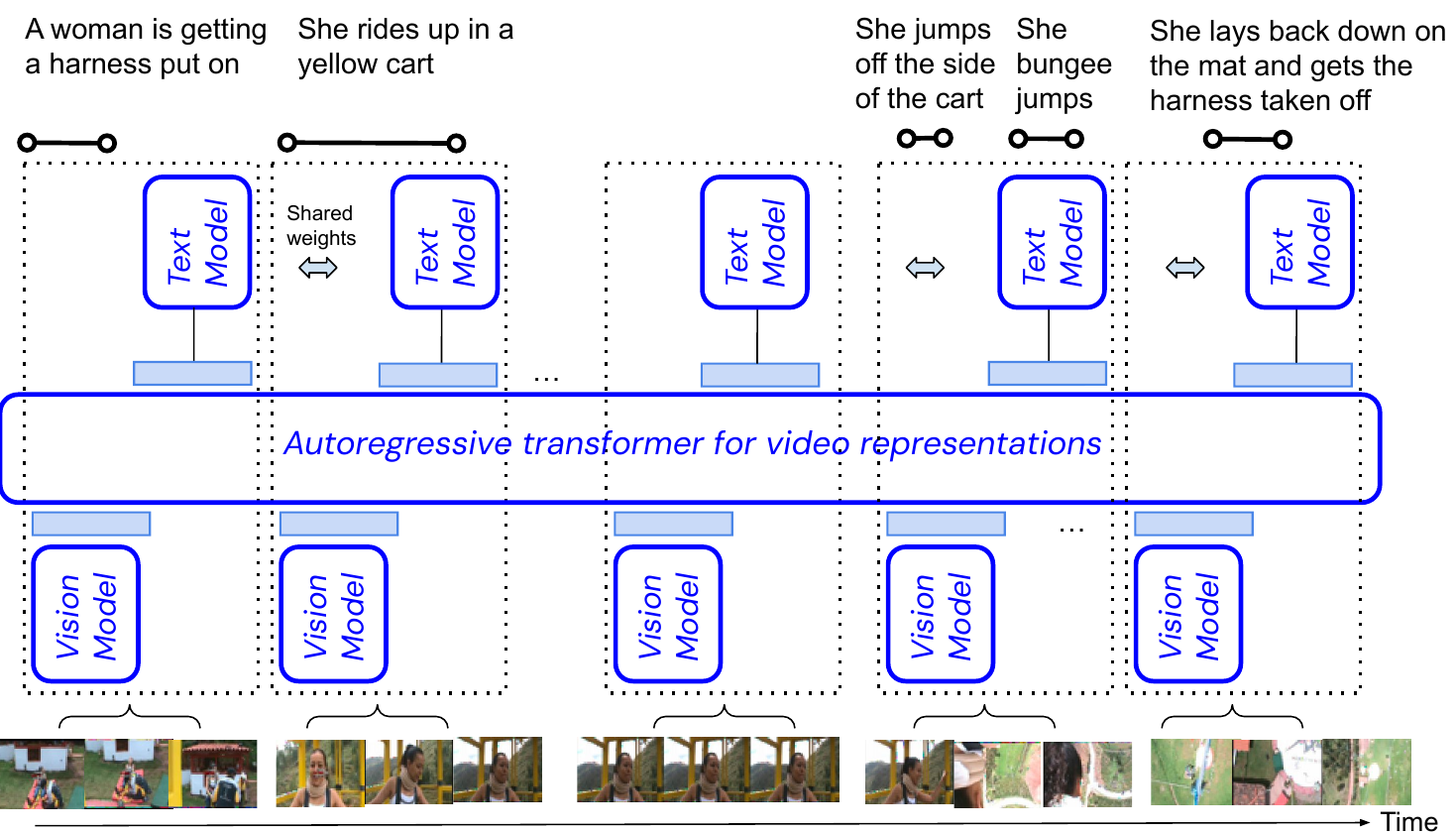}
   \caption{ 
   Our online dense video captioning and event localization model 
   produces rich and granular descriptions in a streaming mode, without access to the future video content.
   %
   %
%
   A key difference to our design is
   the factorized autoregressive decoding, which effectively leverages prior context to generate localized descriptions that are temporally aligned with the video.
   This allows the model to produce comprehensive dense captions, avoiding duplications and including the option to produce more than one output at a time, or no outputs, if applicable.
   %
}
   \label{fig:teaser}
\end{figure}

While understanding videos {\it densely} is one of the most practically relevant tasks, it is also very challenging. Dense video captioning requires both understanding the events in the video in detail, through dense text captioning, and also identifying the start and end timestamps of each of the events. Solving these tasks requires not only understanding of the semantic content, but also parsing of multiple events across the video, understanding when one event starts and finishes and how events are related. This is particularly challenging for long videos because of the large volumes of data, which require efficient modeling, and the long sequence length needed to provide comprehensive captions. 

\begin{figure*}  [t]
    \centering
   
   \includegraphics[width=0.97\linewidth]{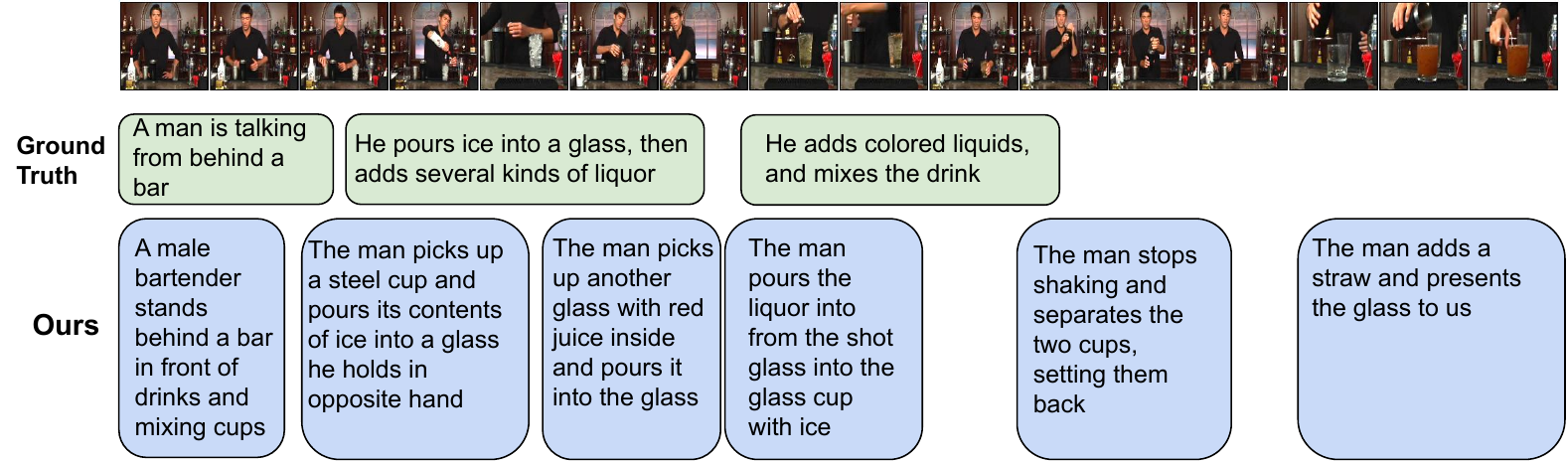}
   \includegraphics[width=0.97\linewidth]{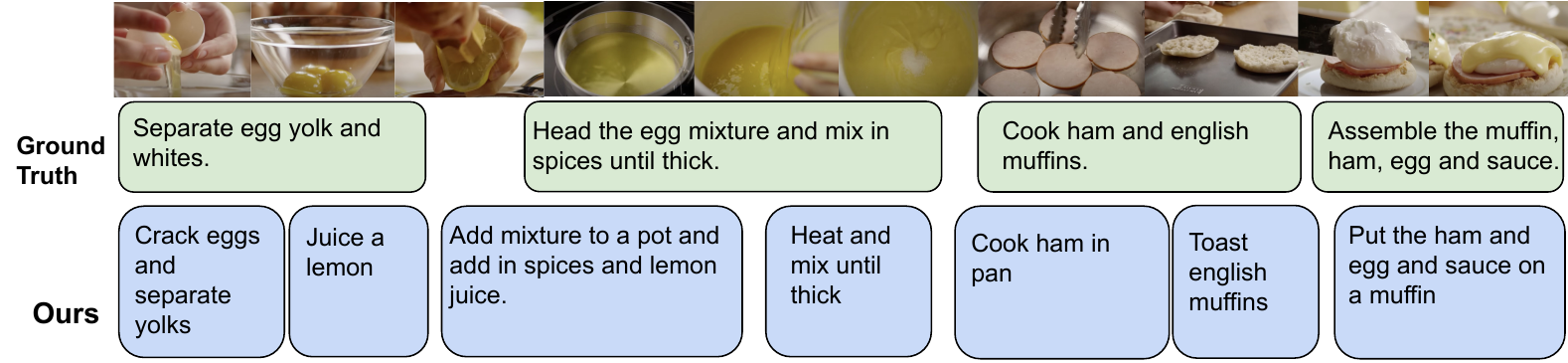}
   \caption{
   The model outputs dense captions for long videos, which are generated much more frequently than the ground truth and are more detailed and specific. 
   The model automatically determines when no output is needed (e.g., when no activities are present or the caption is redundant) and is able to produce outputs in an online fashion, without requiring future video frames, or the entire video.
   }
   \label{fig:visuals}
\end{figure*}

Contemporary Vision-Language Models (VLMs), often built on Large Language Models (LLMs), have advanced video understanding significantly, relying on data and model scaling~\cite{wang2022git,merlot,videoOFA,InternVideo,fu2021violet,mPLUG-2,pali,palix,flamingo}. At the same time, their model structure provides a single text output which is well aligned with ``global'' video understanding tasks such as video captioning, video question-answering or summarization. 
When applied to dense video tasks, e.g.,~\cite{vid2seq,flamingo}
these approaches do not scale very well with increased video length, or the output text length. For example, if we have a several minute long video with tens of thousands of tokens, the attention operation of LLMs will be quite expensive. Furthermore, when increasing the text sequence lengths, e.g., to produce descriptive captions using more words, models using a single LLM will need to process and output many more tokens, further increasing the compute cost.
Lastly, global methods are not applicable in online settings as they require  the full video. 
This is especially limiting for longer videos.

We propose a factorized decoding approach where a single text decoder generates text for {\it multiple localized intervals} within a video, so the compute scales linearly with the video length. This encourages the model to focus on a specific segment and generate a caption about it, making it better able to densely annotate the video and produce more localized and detailed descriptions (Figures~\ref{fig:teaser},~\ref{fig:visuals}).
This allows for more thorough extraction of information from the video, surpassing the limitations of the ground truth. The model also has a memory-like mechanism enabling understanding of the temporal structure and previous intervals of the video, while still operating locally. It maintains an intermediate representation which connects the current information with prior context and is updated with new video frames. 

The proposed model also naturally operates in an online setting i.e., is able to provide a response at various points in the video, and without requiring or `waiting' for future content. It can theoretically work on streaming videos and continue providing outputs even though earlier parts of the video are no longer available in memory. 

Furthermore, we address another common challenge for the dense captioning task, namely dealing with longer videos which also have more output captions and more frames, which can exceed the capacity of current, standard hardware memory. Specifically, we use a cross-segment masking mechanism and design our model to efficiently use a single decoder, invoked multiple times with different inputs alongside the video and thus save parameters and memory during learning. We use learnable, latent video representation that is able to learn both segment-level features as well as leveraging features from prior video segment. This memory-based learning captures the information over much longer durations. Overall, the technical contributions are:
\begin{itemize}
    \item A novel {\it factorized decoding} style architecture for dense video-text tasks. Instead of a global decoder, the model uses a decoder applied to multiple segments and uses a memory mechanism to retain previous segment information. It is able to output multiple captions and event boundaries, at dense intervals (multiple locations) in the video and can scale with the video length.  The results of the model exceed the frequency and detail of the ground truth annotations and provide insights for future data collection efforts for dense captioning tasks.
    \item A decoder-sharing and cross-segment masking mechanism which is applied over the outputs of an autoregressive transformer that captures all the prior context of the video and the current segment details. This results in efficient training and inference for long videos. 
\end{itemize}

\textbf{Results.} 
We evaluate our model on well-established dense video captioning benchmarks: ViTT \cite{vitt}, YouCook2 \cite{youcook2} and ActivityNet \cite{krishna2017dense}, which contain untrimmed long videos. 
We experiment without an ASR input, so as to evaluate the performance of the model from video-only input. The ASR content largely overlaps with the ground truth captions on these datasets, and its occurrence is a strong indicator of actions. Our model outperforms the state-of-the-art approaches in all metrics on ViTT and YouCook datasets, and on SODA and METEOR on ActivityNet, in some cases showing large margins of improvement. The model uses 20\% less compute.

\section{Previous work}

\textbf{Image-Language and Video-Language models.}
Image-Language models, e.g.,~\cite{wang2022git,pali,wang2021simvlm,wang2022image,chen2020uniter,VinVL,tan2019lxmert,li2020oscar,vilbert2020,su2020vlbert,meter,wang2022git,albef,singh2022flava,li2022blip,yu2022coca,wang2022unifying,UnifiedIO,bao2021beit} have demonstrated impressive capabilities to adapt to a broad range of image-text tasks. Furthermore, they often serve as foundational models for video-text tasks. This is done by transfer from image-text to video-text by a number of image-to-video extensions and adaptations~\cite{wang2022git,merlot,dynpretr,palix,videococa,flamingo,MaMMUT}. 
For example, a straightforward solution is to use the frame-by-frame modeling, e.g., GIT~\cite{wang2022git}, MeRLOT~\cite{merlot}, VideoOFA~\cite{videoOFA} used video representation built from frame-level image representations. Flamingo~\cite{flamingo} used a Perceiver resampler~\cite{perceiver} to reduce the volume of the features obtained from videos. Some others, e.g.~\cite{MaMMUT} use learnable sparse video features directly from Tube-ViT~\cite{piergiovanni2022tubevit} for more efficient video adaptations.
Conversely, `video-first' foundational models~\cite{fu2021violet,mPLUG-2,InternVideo} focus on spatio-temporal understanding rather than frame-by-frame intake.
Alternatively one can learn Vector-Quantized video representations 
as in 
~\cite{2023videollm}.
Video-Language models and video-first models alike focus on tasks with short video inputs, reporting results on videos sequences spanning only a few seconds.
Furthermore, some of the most prominent benchmarks consist of video segments which are only 3-10 seconds long~\cite{ucf101,Carreira_2017_CVPR,somethingsomething}.

\textbf{Long-Video modeling.}
Long-form video understanding~\cite{towards-long-form,long-form,kumar2023hiervl,gao2023mist,Tallformer,longmovie,eclipse} has only recently become of interest, where cross-attention~\cite{open-ended} and hierarchical approaches to long videos~\cite{kumar2023hiervl} are popular. 
Some related works approach video understanding by subdividing the video~\cite{CLIP4Clip,CLIP4Caption}. Our model, in contrast, efficiently utilizes the context from video inputs in an autoregressive manner, to extend to much longer videos.
Most above-mentioned approaches focus on video retrieval or video captioning, where a single embedding or caption is produced per video.

\textbf{Dense video understanding.}
Dense video understanding relates to several tasks, for example, event detection, temporal action localization~\cite{tridet,Tallformer,actionformer}, or dense video captioning~\cite{Lioweidensecap,vid2seq,end2enddense,e2edensecap}.
Some approaches propose to detect the segment boundaries first, and then caption in a two-stage manner~\cite{iashin2020abetter}. Others choose to jointly predict boundaries and captions~\cite{wang2018bidir,end2enddense,zhang2022unifying,zala2023hierarchical}. 
PDVC~\cite{end2enddense} uses multiple output heads, one for the temporal interval prediction and one for text. 
Using large language models as foundation and large-scale pretraining, Yang et al.~\cite{vid2seq} and Zhu et al.~\cite{e2edensecap}, formulate the dense video captioning outputs, including start and end timestamps and the caption per segment, as text generative outputs. 
%
%
Longer videos are challenging for these approaches, due to the quadratic scaling of compute for attention mechanisms, 
which greatly increases the compute cost, as videos get longer.
Recent online dense captioning work proposed handling the video history by clustering of previous tokens in a memory module~\cite{zhou2024streaming}.  



\section{Main model}

\subsection{Preliminaries}
\label{sec:prelim}

\textbf{Autoregressive models.} Autoregressive models are commonly applied to sequential data where the goal is to predict the next token in the sequence, given the previous tokens. More specifically, these models introduce a hidden representation $h$, where the input data $x=x_1,\ldots, x_T$ is modeled sequentially, at each step conditioning on the hidden representation for the prior step. That is:
\begin{equation}
p(x_1,\ldots, x_T) = \prod_{t=1}^T p(x_{t+1}|h_t)p(h_t|x_t)
\label{eq:mm}
\end{equation}
Here the input data $x$ is a sequence of length $T$, which, in the case of text, consists of tokens; $h=h_1,\ldots, h_T$ are the hidden representations. The models are trained by a next token prediction loss, e.g., a cross entropy loss for word tokens using a large, predefined dictionary.  

\textbf{Autoregressive modeling of sequential visual features.}
Autoregressive models have shown impressive generative and understanding abilities in LLMs.
A straightforward extension for visual data is to tokenize the visual patches (as in VQ-VAE), i.e., provide a token ID after clustering the data, and treat them in similar way to text tokens in a sequential model. However, with videos the number of visual patches becomes extremely large, needing to span both the spatial and temporal dimensions.

In order to handle long videos, we here use a slightly different approach by still sequentially modeling the features from the video, but instead use a reduced representation for each video segment, rather than using all the patches from the video.
Specifically, we simply subdivide the video frames into short non-overlapping segments~\cite{MoCha,CLIP4Clip}, extracting visual features per video segment. 
The output of the autoregressive model for segment at step $t$ are thus a learned combination of all the previous features and the current features. This allows the autoregressive model to act as a form of memory while reducing the compute needs due to the smaller segment level representations. 

\subsection{Model architecture}

At a high level our model can be thought of multiple VLM-like models that process the video locally, per segment (Figure~\ref{fig:teaser}). By processing short segments, this allows the model to handle longer videos without significantly increasing compute requirements, due to the linear scaling of compute cost with respect to the number of segments, rather than the quadratic costs associated with increasing sequence length for self-attention. Several innovations of this work allow the model to do that: First, due to the autoregressive model representation abilities and memory mechanisms, the model is able to pass compressed information from previous segments to future ones, allowing the text decoder to know what happened previously when generating a caption for the current segment. Second, we propose a masking mechanism which allows the decoder to access various inputs without having to re-compute them in memory, which leads to efficient training and inference. Third, we note that we share the weights of the video model and the text model for each segment to maintain a small model size.

\begin{figure*}
    \centering
    \includegraphics[width=0.95\linewidth]{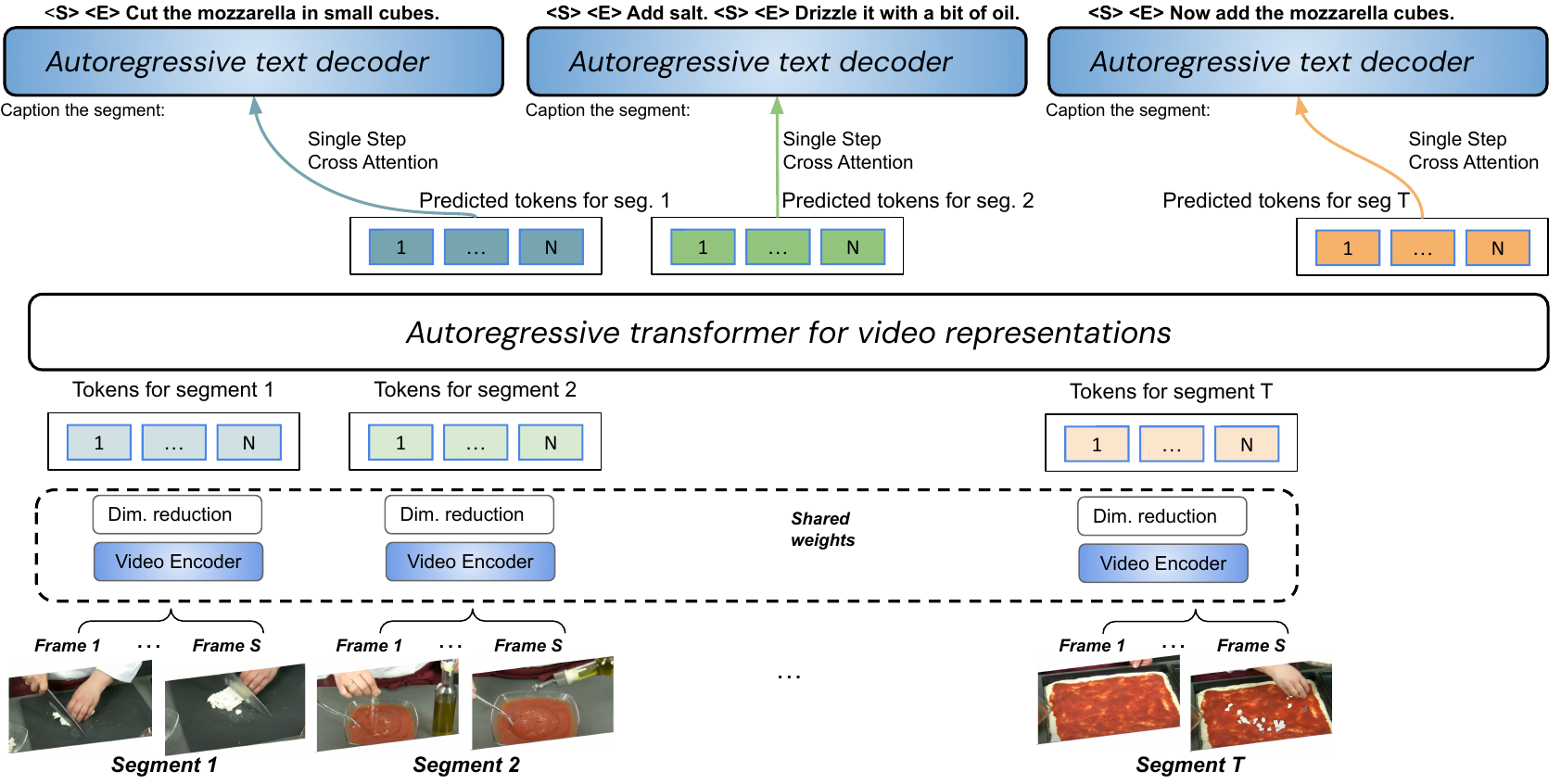}
   \caption{Model architecture overview. The model consists of multiple decoders which are responsible for captioning video segments. The autoregressive transformer models the video temporally, conditioned on the previous feature representation, providing a higher level of abstraction and a longer-range temporal modeling. Each local decoder is able to ``see'' information of features before the current segment, thus understanding the current events in context to prior ones. This allows a more detailed and localized descriptions per video corresponding with where the activities occur in the video. 
   The required output format is `start` of segment token \textit{$<$S$>$},  `end` token \textit{$<$E$>$} and a caption. The text shown at the top per decoder is provided only during training.}
   \label{fig:main}
\end{figure*}

Our model, visualized in Figure~\ref{fig:main} consists of the following components: a video encoder, dimensionality reduction module, autoregressive memory component to learn temporal structure, and the factorized text decoder. Importantly, this partitioning of the video and dense decoding allows the model to more descriptively caption local segments and scale to longer videos when compared to previous approaches.  We describe these components in detail below.

\textbf{Video Encoder.}
First, given a video $v$ which consists of $L$ frames, we first split this video into $T$ segments of $S$ frames (i.e., $S=\cfrac{L}{T}$, $T$ is a hyper-parameter). These segments are then passed to a TubeViT~\cite{piergiovanni2022tubevit} model, $\phi$ to create a set of features for the video segment, $r_v=\phi(v)$, outputting a $[T, K]$ tensor. We choose to use TubeViT as it showed strong results in previous works while also reducing the number of features for a video, which reduces the cost of self-attention throughout the model.

\textbf{Dimensionality Reduction.}
While the video features per segment are already much smaller than with other approaches, we also use a transformer model, $X$ to further reduce the dimensionality, similar to \cite{mirasol3B}. Inspired by prior work, we use a transformer taking in the $K$ tokens for this segment, and using the last $N$, $N<<K$, tokens as the reduced representation as output for this segment, $r=X(r_v)$, outputting a $[T, N]$ tensor. Due to the self-attention in this transformer, it can learn to merge all the features into the last $N$ ones. This is similar to other attention-based mechanisms which demonstrated that transformers can effectively reduce the dimensionality~\cite{ttm,bulatov2022recurrent,darcet2023vision,ryoo2021tokenlearner_neurips,perceiver}.

\textbf{Autoregressive Transformer.}
Since the segments only encode features from a short video snippet, we next use an autoregressive transformer, $\psi$ to learn temporal structure over the longer duration of the video. Since the video encoder and the dimensionality reduction transformer have already `compressed' the representation to a shorter sequence of features, this autoregressive model can learn longer temporal relationships over the video representations without significant compute cost. The autoregressive transformer functions as described in Sec.~\ref{sec:prelim}, i.e., a standard transformer model with causal masking applied over the segments, $m=\psi(r)$, outputting a $[T, N]$ tensor, same dimensions as the input. This allows the output of the autoregressive model for a segment to contain information from the prior segments, functioning as a memory mechanism.

\subsubsection{Factorized Text Decoding}
A standard approach of reusing trained LLMs/VLMs is to apply the text decoder on top of the video or image tokens. Previous work have done this successfully, e.g.,~\cite{vid2seq,flamingo}. However these approaches do not scale very well with the increase of video length, or required output length,
As video gets longer, the compute costs increased significantly, due to the quadratic scaling of compute of attention mechanisms. Furthermore, when increasing the text sequence lengths, e.g., more descriptive captions using more words, models using a single LLM will need to process and output many more tokens, further increasing their compute cost.

We here instead propose {\it factorized decoding} where a text decoder, $D$ generates a caption for each video segment $t$ independently. I.e., $o_t = D(m_t)$, the decoder is independently applied to each segment, so compute linearly scales with video length. 
This also enables the model to focus on a specific segment and generate a caption locally and according to the video content.
Thus the model is able to densely annotate the video and produce more local and detailed information, rather than mimicking the ground truth which might be missing (Figure~\ref{fig:visuals}).
Our text decoder takes the visual features of a segment as inputs to a cross-attention layer, and outputs a short, local caption, rather than a longer global caption. This setup reduces the cross-attention size and output length per segment, greatly reducing compute costs. The shared decoder can be pre-trained on other video-language tasks.

Additionally, we design the factorized decoders to leverage all the context information in an efficient way. Since the autoregressive transformer learns temporal information, the cross-attention input features have access to all the prior video features, allowing the dense decoders to integrate information from the earlier parts of the video, rather than just the current segment. This lets the model understand the structure of a long video and generate good contextualized captions, rather than only independent captions. Importantly, since the text decoder is only running on a segment, rather than the whole video, this lets the video inputs be a smaller sequence of features, rather than all the features from the entire video. This greatly reduces the compute requirements of the model and improves scaling to long videos. Further, this allows for more specific, local captions about a segment of the video, rather than having to generate a long caption describing the entire video.


\subsubsection{Cross-segment masking}
\begin{figure} 
    \centering
    \includegraphics[width=0.95\linewidth]{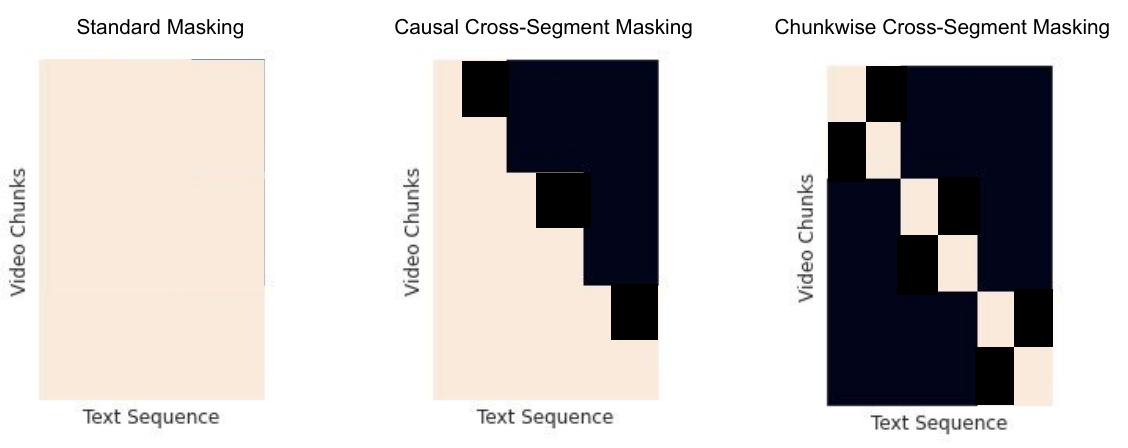}
    \caption{Example of standard, global cross-segment masks, and the causal and segment-wise mask we explore for training here.}
    \label{fig:xattnmask}
\end{figure}

Furthermore, we designed an efficient implementation of the model. During training, rather than running the text decoder $T$ times, i.e., once per segment, which we found to increase memory usage, we instead run the decoder once to generate the full sequence. We mask the cross-attention so the decoder for segment $s$ can only see the associated video features for that segment. I.e., rather than generating $T$ captions of length $l$, we run the decoder once, generating a sequence of length $T\cdot l$. However, we create a mask for cross-attention that ensures that the generation at time step $s$ ($s$ between 0 and $T\cdot l$) can only access the video features associated with segment $s$ ($s=\lfloor\frac{t}{l}\rfloor$). We pad all sequences to be $l$. This mask, visualized in Figure \ref{fig:xattnmask} (right), shows the cross-attention features that can be used at each text sequence location. For example, the leftmost one shows standard cross-attention where any feature can be used at any text location.
The middle one shows a causal segmented version of the masking and the rightmost one shows our masking where each segment of text can only attend to the associated segment of vision features, relying on the autoregressive model to capture temporal information. We note that during training, this also saves memory. 
Importantly, this leads to realized compute savings  of 20\% during inference, as we observed in our experiments. 

During inference, to fully realize the compute savings, we can apply the decoder once per segment, in a streaming setting, using only the segment video features as cross-attention inputs. In inference, we do not need to save the activations for back-prop, so this does not increase memory usage and enables fast inference.

\subsection{Application to dense captioning}
To apply our model to the localization setting, e.g., dense captioning, we use the text format proposed in Vid2Seq \cite{vid2seq}. Specifically, we represent a localization caption as a string with the form: \textit{$<$start$\_$token$>$ $<$start\_time$>$ $<$end\_time$>$ $<$caption\_text$>$ $<$EOS$>$}. The start token is discretized. This allows representing a localized caption or action as tokens, rather than requiring multiple output heads to generate times and text. We also experiment with the discretization settings. Different from Vid2Seq, we here further align the captions with the video segments. Specifically, during training, we make the labels for a segment be the text with an end time within the segments temporal interval. If a segment has no actions, we simply annotate that segment with \textit{$<$EOS$>$}. If it has multiple actions, we combine them: \textit{$<$start$\_$token$>$ $<$start$\_$time 1$>$ $<$end$\_$time 1$>$ $<$caption$\_$text 1$>$ $<$start\_token$>$ $<$start\_time 2$>$ $<$end\_time 2$>$ $<$caption\_text 2$>$ ... $<$EOS$>$}. This allows the model to localize captions to segments as it sees them, rather than having to generate all the localization at the end of the video. This alignment can be noisy if segments do not align well to the captions, but the model tends to follow the ground truth in assigning the caption to only one segment and has no issue with more than one segment. We found this simple approach to work well.

During inference, we apply beam search to generate the captions, then use temporal NMS to remove any intervals with greater than 0.7 temporal IoU (i.e., remove any generated overlapping captions). The remaining intervals and their predicted texts are used to compute the metrics.

\section{Experiments}

\begin{table*}[t]
\centering
{\fontsize{8pt}{8pt}\selectfont 
\begin{tabular}{l|c|cccc|cccc|cccc}
& 
&  \multicolumn{4}{c|}{ViTT} & \multicolumn{4}{c|}{YouCook2} & \multicolumn{4}{c}{ActivityNet} \\
 Method  & Pretraining & S    & C     & M    & F1     & S   & C   & M   & F1     & S   & C   & M   & F1   \\
\midrule


E2ESG~\scriptsize{\citep{e2edensecap}}  & No
    & - & - & - & -     & - & 25.0 & 3.5 & -    & - & - & - & -    \\
MT~\scriptsize{\citep{Lioweidensecap}}  & No
  & - & - & - & -     & - & 6.1 & 3.2 & -    & - & 9.3 & 5.0 & -    \\
PDVC~\scriptsize{\citep{end2enddense}}  & No
    & - & - & - & -     & 4.9 & 28.9 & 5.7 & -    & 6.0 & 29.3 & 7.6 & -    \\

TimeChat~\scriptsize{\citep{Timechat}}  & No
     & - & - & - & -      & 3.4  & 11.0 & - & 19.5    & - & - & - & -  \\

GIT~\scriptsize{\citep{wang2022git}}  & Multiple\citep{wang2022git}*
      & 7.1 & 15.1 & 3.4 & 32.5     & 3.1 & 12.1 & 3.4 & 17.7    & 5.7 & 29.8 & 7.8 & 50.6    \\

OmniViD~\scriptsize{\citep{wang2024omnivid}}  & Kinetics-400
     & - & - & - & -       & - & - & - & -     & - & 26.0 & 7.5 & - \\

Vid2Seq $\dagger$~\scriptsize{\citep{vid2seq}} & YT-Temporal-1B
    & 9.8 & 23.0 & 5.0 & 37.7      & 5.7 & 25.3 & 6.4 & 23.5    & 5.9 & 30.2 & 8.5 & 51.8    \\

DIBS~\scriptsize{\citep{wu2024dibs}} & Custom HowTo\citep{wu2024dibs}*
  & - & - & - & -      & 6.4 & 44.4 & 7.5 & 31.4    & 5.9 & 31.9 & 8.9 & \textbf{55.6}    \\
%
Zhou et al.~\scriptsize{\citep{zhou2024streaming}} & WebLI~\cite{pali}*
    & 10.0 & 25.2 & 5.8 & 35.4      & 6.0 & 32.9 & 7.1 & 24.1    & 6.2 & \textbf{37.8} & 10.0 & 52.9    \\
\midrule


\textbf{Ours} & HowTo  
   & \textbf{10.2}    &\textbf{37.2} & \textbf{18.9} &\textbf{47.4}  
   & \textbf{11.3} & \textbf{57.6}  & \textbf{27.7} &\textbf{33.6}
   & \textbf{12.3} &18.4 &\textbf{19.5} & 53.3  \\
\bottomrule
\end{tabular}
}
\caption{
\textbf{Dense Captioning and Event Localization results for VITT, YouCook2 and ActivityNet datasets.}  Reporting SODA (S), CIDEr (C), METEOR (M) scores, and localization metrics (F1). Our model performs well, even without pre-training. 
It has an additional benefit of being an online method, similar to Zhou et al.~\cite{zhou2024streaming} (shown at the bottom. 
Results which were not reported in previous work are shown as `-'. 
$\dagger$: version with visual-only inputs. 
*Datasets are either proprietary or non-public processing is applied.
We use video-only inputs as majority of the SOTA approaches. 
}
\label{tab:sota}
\end{table*}

We conduct experiments on standard, challenging datasets for dense video captioning: ActivityNet dense captions \cite{krishna2017dense}, ViTT \cite{vitt} and YouCook2 \cite{youcook2}. The goal of dense video captioning is to output multiple captions and associated start and end times of those captions. 
This implicitly requires the model to both segment the video into meaningful events and provide captions corresponding to the events.
We compare the standard SODA \cite{fujita2020soda}, CIDEr \cite{vedantam2015cider} and METEOR \cite{banerjee2005meteor} scores as done in prior works~\cite{vid2seq}. We also report the 
F1 score which is based on the IoU (Intersection over Union) of the predicted temporal intervals with the ground truth events 
for these datasets. We note that evaluation for these tasks is in itself challenging. The descriptions are free-form and subjective, and it is not clear when a segment starts and finishes. Because of this, the evaluation metrics are not always indicative of performance. As a result, the evaluation metrics will penalize approaches like ours which can output many more detailed, but still accurate, descriptions, while the ground truth is provided more sparsely.

\textbf{Datasets.}
The \textbf{VITT} dataset~\cite{vitt} is a dense video captioning dataset of instructional videos.  The videos are around 250s long on average. 
The dataset consists of 8K untrimmed instructional videos and has around 7.1 temporally-localized annotations per video.
\textbf{ActivityNet
Dense Captions}~\cite{krishna2017dense} is a video dataset over long videos containing a variety of human activities. The videos are 120s long on average. The dataset has 20k untrimmed videos, where each video annotated with an average of 3.7 localized captions.
\textbf{YouCook2} Dense Captions~\cite{youcook2} has 2k untrimmed, long cooking videos. The videos are about 200s length on average with annotated with 8 localized captions, on average.
\textbf{HowTo} We also use HowTo100M \cite{HowTo100M} for pretraining, following the same settings as the other datasets, but using the time-stamped ASR as the target dense captions, which is a rather noisy training signal.

\textbf{Model size.} Our model has about 500M parameters -- 128M are for the text decoder, 300M for the vision encoder (TubeViT with ViT-Large), and the remaining parameters are for the dimensionality reduction transformer and autoregressive transformer, about 32M each. 

\subsection{Experimental results}
Table~\ref{tab:sota} shows the results of dense video captioning on all three datasets, VITT, ActivityNet and YouCook2, comparing to the state-of-the-art. We report the standard metrics for dense captioning such as METEOR \cite{banerjee2005meteor}, CIDEr \cite{vedantam2015cider}, SODA \cite{fujita2020soda} and F1. 
Here we only input a video to the model and require it to output the localized captions and their extents.
We note that the prior Vid2Seq~\cite{vid2seq} work additionally uses timestamped ASR as input to the model, so we only report its vision-only performance.  Our setting is more challenging and realistic, as the additional timestamped ASR often overlaps with the ground truth captions. 

Table~\ref{tab:sota} (left) shows that the model outperforms the state-of-the-art on the VITT dataset, on all metrics and in the cases of CIDEr and METEOR, by large margins. 
%
Table~\ref{tab:sota} further measures the event localization performance. As seen, our model outperforms all state-of-the-art approaches with large gains on F1 score, as well. 
Comparing to the other online approach~\cite{zhou2024streaming}, we also see that our model outperforms it.
All the other approaches are offline, which means they require the full video in order to output results.




Table~\ref{tab:sota} (middle) further compares the performance on the YouCook2 dense captioning benchmark. We also see here too that our model outperforms all state-of-the-art approaches and also by large margins.

Table~\ref{tab:sota} (right) compares the performance on ActivityNet. We see  large improvements over prior work on SODA and METOR. 
The CIDEr score of our method tends to be lower. CIDEr is generally more unstable and may not be fully reliable, as noted in previous work \cite{cider_unstable}. 

For localization, we find that our F1 scores are overperforming the state-of-the-art, in many cases by large margins, with the exception of ActivityNet, where the DIBS model~\cite{wu2024dibs} has the highest value. We note that they also used custom re-captioned HowTo100M dataset, whereas we used the original one based on noisy ASR labels.
We also observe that our model performs well even without pre-training (please see supp. material for more results). The YouCook2 datasets seems to benefit the most from pre-training, likely due to it being a small dataset (2000 videos).

Moreover, we observe that our model  tends to give more frequent but still meaningful descriptions (Fig.~\ref{fig:visuals}, also please see the supp. material for statistics), whereas the ground truth annotations are more sparse, as providing dense annotations is costly. This might cause mismatch and subsequently lower metrics as the predicted outputs are deemed as not matching sufficiently to the ground truth. Despite that, having denser caption outputs is much more useful in practice.
This observation provides further insights in future data collection. As seen, \textbf{it is now possible to provide much more dense outputs by an automatic model}. This will allow the model to provide more dense data labeling, which is otherwise a costly and tedious process. 

\subsection{Model architecture efficiency}

In Table~\ref{tab:flops} we explore the efficiency of the proposed factorized architecture, comparing to the global one, when both are on equal footing.  
We find that for our standard setting, using 512 frames and 8 or 16 segments and 32 output tokens per segment, the model uses 424 GFLOPs per segment. For 8 segments, this is 3392 GFLOPs (total of 256 output tokens) and for 16 segments, 6784 (512 output tokens). In contrast, the global decoder baseline model, which has the same model components and number of parameters, but does not factorize the decoder, uses 4125 GFLops for 256 output tokens and 8462 GFLOPs for 512 output tokens. This is a \textbf{18-20\% savings in FLOPs} and the \textbf{FLOPS reduction increases} as video and caption length increases, due to the quadratic cost of attention.

\subsection{Ablations}
We conduct ablations to study the various components of the model. To minimize compute usage, we use lower spatial resolution (224$\times$224) and 128 frames for the ablations, unless otherwise noted. 

In Table \ref{tab:model-ablation}, we compare the effects of the components of the model. The baseline is a single, global text decoder model taking all video features as cross-attention inputs. We find that factorizing the decoder, adding dimensionality reduction, the autoregressive module and all components together all help improve results. The differences are small, but show some consistent improvement across the metrics by adding them. Notably, we find that the factorized decoder improves on all metrics and, in addition, has meaningful compute savings (Table \ref{tab:flops}).  



Table~\ref{tab:vs-baseline} shows the comparison of the proposed dense decoding model vs a single-decoder model. We use 512 frames to further show the benefit of the dense decoder. The single decoder is similar to~\cite{vid2seq}, however we do not use ASR for either experiment. We see that our model is outperforming the single-decoder. It also has the advantage that it can scale to much longer videos with more outputs.

Table~\ref{tab:frame-chunk-ablation} 
studies the effect of the number of segments and frames on the overall performance. The number of segments determines the number of decoding outputs (where each segment can output 0 or more captions). 
We see that, in general, more frames is beneficial, as expected. Also we see that here matching the number of segments to the ground truth works better which is not surprising. Note that VITT averages 7 descriptions per video.




\begin{table} [] 
\centering
\begin{tabular}{c|ccc}
Model	& Global & Factorized & Savings (\%)  \\
\midrule
8 seg, 256 tok & 4125 & 3392 & \textbf{+18\%} \\
16 seg, 512 tok  & 8462 & 6784 & \textbf{+20\%} \\
\bottomrule
\end{tabular}
\caption{\textbf{GFLOPs savings.} Comparison of GFLOPs of our Factorized model vs the Global model. Savings of 18-20\% are achieved.}
\label{tab:flops}
\end{table}

\begin{table} [] 
\centering
\begin{tabular}{l|ccc}
Model	& SODA & CIDEr & METEOR  \\
\midrule
Base & 4.0 & 10.4 & 5.0 \\
+ Factorized Decoder & 4.2 & 11.7 & 5.1 \\
+ Dim Reduction & 4.1 & 11.8 & 5.2 \\
+ Autoregressive & 4.1 & 11.8 & 5.1 \\
+ All (full model) & 4.2 & 12.0 & 5.2 \\
\bottomrule
\end{tabular}
\caption{\textbf{Model components.} All components are helpful. 
The best performance is when all of them are included together.}
\label{tab:model-ablation}
\end{table}

\begin{table} [] 
\centering
\small
\begin{tabular}{l|ccc}
	& SODA & CIDEr & METEOR  \\
\midrule
Global &4.8 & 16.8 & 7.7 \\
Dense (Ours) &6.5 &26.0 &8.2 \\ 
\bottomrule
\end{tabular}
\caption{\textbf{Dense vs Global single decoder.} The dense decoder outperforms the single global decoder. Uses 512 frames. ViTT dataset.}
\label{tab:vs-baseline}

\end{table}

\begin{table} [] 
\centering
\begin{tabular}{l|ccc}
Frames/Segments	& SODA & CIDEr & METEOR  \\
\midrule
32/8 & 4.3  & 11.9 & 5.4 \\
64/8 & 4.2  & 11.9 & 5.3 \\
128/8 & 4.1 & 12.0  & 5.4 \\
256/8 & 4.4 & 11.8 & 5.2 \\
512/8 & 4.5 & 13.3 & 5.3 \\
256/16 & 4.1 & 12.7  & 5.2  \\
512/16 & 4.2  & 12.8 & 5.1 \\
\bottomrule
\end{tabular}
\caption{\textbf{Sensitivity to segments.} Studying number of segments, which control the number of decoding outputs on the ViTT dataset.}
\label{tab:frame-chunk-ablation}
\end{table}

\begin{table}
\centering
\begin{tabular}{l|ccc}
	& SODA & CIDEr & METEOR  \\
\midrule
Absolute & 4.1  & 11.9 & 5.2 \\
Relative & 4.2  & 12.0  & 5.2 \\
\bottomrule
\end{tabular}
\caption{\textbf{Absolute vs relative time} Comparing time that is absolute vs. relative to the length of the video, on ViTT.}
\label{tab:time-setting}
\end{table}

\begin{table}
\centering
\begin{tabular}{l|ccc}
Format	& SODA & CIDEr & METEOR  \\
\midrule
start-end &  4.2 & 12.0 & 5.2 \\
center-duration & 3.9  & 11.6 & 5.0 \\
\bottomrule
\end{tabular}
\caption{\textbf{Time format.} start-end vs. center-duration format of the time.}
\label{tab:format}
\end{table}

\begin{table}
\centering
\begin{tabular}{l|ccc}
\# Time Bins	& SODA & CIDEr & METEOR  \\
\midrule
32 &  4.9  & 22.4 & 6.7 \\
64 &  4.4 & 15.4 & 5.7 \\
128 & 4.2 & 12.0  & 5.2 \\
\bottomrule
\end{tabular}
\caption{\textbf{Number of bins.} Studying of bins we discretize the time into, on ViTT.}
\label{tab:time-bins}
\end{table}

\begin{table}
\centering
\begin{tabular}{l|ccc}
Prefix	& SODA & CIDEr & METEOR  \\
\midrule
No & 4.2  & 12.0 & 5.2 \\
Yes & 5.1  & 14.2 & 6.7 \\
\bottomrule
\end{tabular}
\caption{\textbf{Use time prefix.} Whether or not a time prefix is added to each chunk, on ViTT.}
\label{tab:prefixed}
\end{table}




We study how the method creates discrete time tokens. Following Vid2Seq, we discretize the time into a set of buckets.
 In Table \ref{tab:time-setting}, we compare using absolute time vs. the time relative to the duration of the video, and in Table \ref{tab:format}, we compare using the start and end times of an interval to the center and duration of an interval. We find relative and start-end format to be better.
In Table \ref{tab:time-bins}, we compare the number of bins used to discretize the time into, finding 32 is better than the 128 previously used. In Table \ref{tab:prefixed}, we find that for our dense model, adding a time token to the prefix of the current segment time, e.g., `Caption the segment: $<$start$\_$time$>$' helps a lot, since this gives the language model a more explicit signal to where in the video it currently is. 
We also study the inference settings of the model: the number of samples generated during decoding (e.g., by beam search), the temperature used, etc., as well as the IoU threshold for temporal NMS. Overall, we found that for these datasets generating 18 samples with a temperature of 1 and an IoU threshold of 0.7 worked best, but these settings are likely specific to these datasets.

\textbf{Visualizations.} Figure~\ref{fig:visuals} shows an example of our model output compared to the ground truth. As seen, the model provides dense captions that are more frequent than the ground truth. The descriptions are detailed and capture more activities in the video. The model does not  output predictions when not needed (i.e., we use 16 segments, whereas in the example we see 6 predictions). However on these datasets, the ground truth captions are fewer and shorter, so the metrics will penalize our method.
Our model excels at providing dense and descriptive captions, particularly useful for long videos.



\section{Conclusions}

We propose a factorized decoding architecture for online dense video captioning and event localization, which is efficient and provides  more dense, descriptive and localized captions. The proposed model both effectively incorporates context from prior segments for comprehensive and non-repetitive outputs  
and, at the same time, saves compute by factorizing the decoder. 
Our approach also provides denser descriptions in videos, adapting to higher frequency of descriptions, especially needed for long videos.

%



\appendix

\section{Additional experiments}

\subsection{Dense Captioning Statistics}

We examine the statistics of the outputs of the proposed factorized autoregressive model. More specifically, we compare them to the outputs of a global autoregressive model couterpart. In Table \ref{tab:numcaptions}, we compare the average number of generated captions and the total number of words generated by the baseline, which is a single global caption model, compared to our factorized dense decoder model. As seen, our model provides more captions per videos and the captions are more detailed using more words on average. 

We observe that our model also provides denser and more detailed descriptions than the ground truth. This is not surprising as manual labeling is tedious and the ground truth captions are generally sparser and do not always cover the video comprehensively (Figure~2 of the main paper). 
As also seen in figure the extra captions of the proposed model provide relevant and detailed information about the video.

\begin{table*}[t]
    \centering
    \begin{tabular}{l|cccc}
    \toprule
          &  \# Predicted Captions & \# Pred Words & \# GT Captions & \# GT Words \\
    \midrule
    \multicolumn{4}{l}{ViTT} \\
    \midrule
    Single, global decoder  & 5.5 & 25.2 &  7.1 & 22.0\\
    Dense Decoder (Ours)  & 12.4 & 52.4 & 7.1 & 22.0\\
    \midrule
    \multicolumn{4}{l}{YouCook2} \\
    \midrule
    Single, global decoder   & 6.8 & 67.5 &  8 & 70.4\\
    Dense Decoder (Ours)  & 15.4 & 103.5 & 8 & 70.4\\
    \bottomrule
    \end{tabular}
    \vspace{1mm}
    \caption{Comparison of the average number of captions and words per caption, generated by our model, compared to a global captioning model baseline and also to the ground truth caption statistics provided in the VITT and YouCook2 datasets. We see that our model, on both YouCook2 and ViTT generates more captions and more words per caption. 
    This indicates our model is able to provide more detailed and frequent captions.}
    \label{tab:numcaptions}
\end{table*}

\subsection{Additional Pre-training results}

Table~\ref{tab:pretr} provides additional results of our approach when no video pre-training is used, compared to training with HowTo100M dataset pre-training, provided in the paper. As seen, pre-training is helpful in most cases, and provides a small improvements in performance. The YouCook2 dataset seems to gain most from pre-training, which is expected as it is a small dataset.
We observe too that our model performs excellently even without pre-training.

\begin{table} [] 
\centering
\begin{tabular}{l|cccc}
Model	& SODA & CIDEr & METEOR &F1  \\
\midrule

\textbf{ViTT}  \\
\midrule

No pre-training   & 9.8  &\textbf{42.2} & \textbf{22.8} & 47.2  \\
HowTo pre-training   & \textbf{10.2}    &37.2 & 18.9 &\textbf{47.4}  \\
\midrule
\textbf{YouCook2} \\
\midrule

No pre-training  & 10.7 & 55.3 & 21.0 & 27.6 \\
HowTo pre-training    & \textbf{11.3} & \textbf{57.6}  & \textbf{27.7} &\textbf{33.6} \\
\midrule
\textbf{ActivityNet}  \\
\midrule
No pre-training & 10.0 & 17.9 & 15.8 & 51.6    \\
HowTo pre-training    & \textbf{12.3} &\textbf{18.4} &\textbf{19.5} & \textbf{53.3}  \\
   
\bottomrule
\end{tabular}
\caption{Performance comparison on all datasets, with and without HowTo100M video pre-training.}
\label{tab:pretr}
\end{table}

\section{Additional visualizations}

In Figure~\ref{fig:visuals} we show additional visualizations of the model. As in Figure 2 of the main paper shows, we can see here too that the model outputs accurate captions which are also more frequent and specific.

\begin{figure*} []
    \centering
   \includegraphics[width=1.0\linewidth]{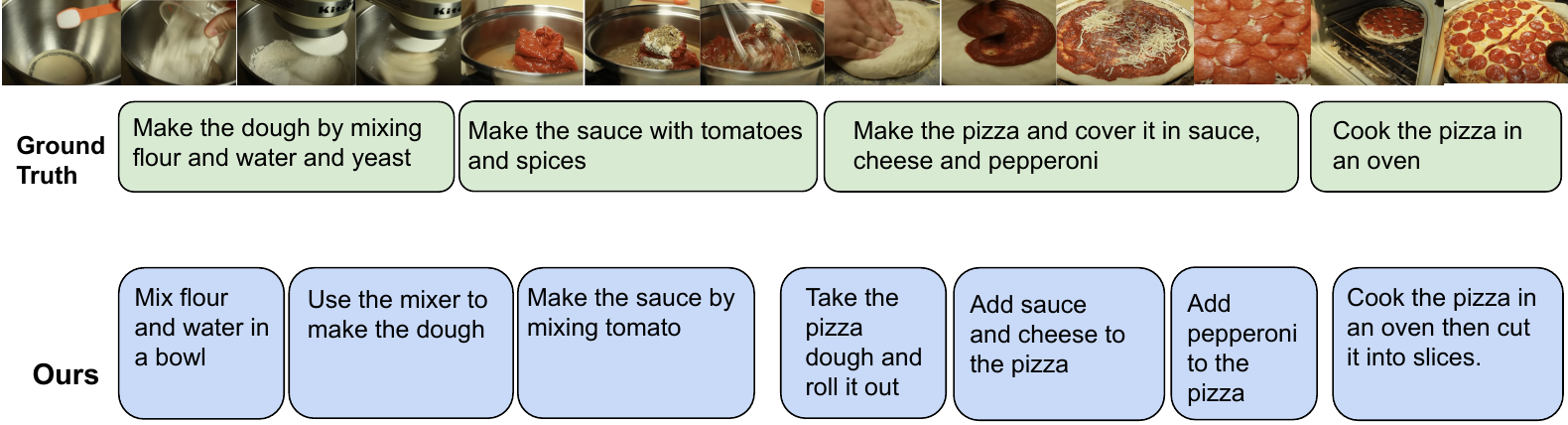}
   \caption{Visualizations of the model outputs. Top: samples from the video sequence. Middle: The ground truth. Bottom: Our model predictions. 
   }
   \label{fig:visuals}
\end{figure*}

\section{Additional implementation details}
The model has about 500M parameters. Of these, 128M are devoted to the text decoder which is shared for each segment. We use 300M parameters for the vision encoder (TubeViT, specifically using a ViT-Large backbone), which is also shared for each segment, and the remaining parameters are for the video memory transformer and autoregressive transformer. Our model used video input of $512$ frames at $448\times 448$ spatial resolution with 16 segments. Due to the compression within the model, even with a high number of frames and high resolution, this fits on 64 devices.

We trained for up to 20,000 steps with a batch size of 16. We used a learning rate of 0.0001 with the Adam optimizer. We used label smoothing of 0.1, dropout of 0.1 and weight decay of 0.00001. For decoding, we used beam search with 24 outputs, followed by temporal NMS with a threshold of 0.7.

The ViT and language model are pre-trained on the ALIGN dataset \cite{align}, using the weights from the MaMMUT pre-trained model \cite{MaMMUT}.

\section{Limitations and broader impacts}
We here discuss the limitations of our model. 

First, it still relies somewhat on the descriptiveness and details in the ground truth captions. For example, as seen in Table \ref{tab:numcaptions}, when trained on short captions as in ViTT, the model tends to produce short captions, but when trained on long captions (e.g. in YouCook2), it generates longer captions. Another limitation of the model is that the number of segments, e.g., 16 in most of our experiments, can influence the number of prediction captions. The model often generates 1 caption per segment, although it is capable of generating more or fewer captions per segment. This is mainly tied to the ground truth, e.g., on ViTT which has fewer captions, the model generates fewer captions, using more empty segments. Thus a very sparsely labeled dataset will influence the model. Another issue is that there is a mismatch between the model and ground truth annotations, as our model outputs more captions and describes more intervals than are annotated in the ground truth. So even if the model is correct, the metrics will penalize the model.

The video dense captioning task is, to an extent, a subjective task, leading to unclear boundaries and mismatches between model outputs and the metrics computed on the ground truth data. 

The model is created for research purposes and its intent is to evaluate its performance in comparison with the state-of-the-art. It demonstrates new capabilities which can have positive impacts. The model is not intended for other uses other than research and inspiring more research ideas.

{
    \small
    \bibliographystyle{ieeenat_fullname}
    \bibliography{main}
}


\end{document}